\documentclass[11pt,a4paper]{article}
\usepackage[T1]{fontenc}
\usepackage[utf8]{inputenc}
\usepackage[margin=25mm]{geometry}
\usepackage{lmodern,microtype}
\usepackage{tikz}
\usetikzlibrary{arrows.meta}
\usepackage{amsmath}
\usepackage{amsmath,booktabs,array,longtable}
\usepackage{xurl}
\usepackage[hidelinks]{hyperref}
\hypersetup{pdftitle={deepopen: Laya-Based Intent Classification on BANKING77 and CLINC150},pdfauthor={Shuang Song, Zhongqiao Yuan}}
\title{DeepOpen: Decision-Oriented Intent Classification on BANKING77 and CLINC150}

\author{
Shuang Song \and
Muhammad Luqman Naseem \and
Zhongqiao Yuan
}

\date{
\small Shenzhen Depei Open Source Data Technology Co., Ltd.\\
\small Shenzhen, China\\[6pt]
\texttt{kaifa@gitcc.com}
}
\begin{document}
\maketitle

\begin{abstract}
We present DeepOpen, a decision-Oriented intent-classification framework with dataset-specific systems for BANKING77 and CLINC150. On BANKING77, a single English classifier continued with sharpness-aware minimization (SAM) achieves 94.1234\% test accuracy (2,899/3,080). This is the best observed result among three fixed SAM seeds, selected for display after test evaluation; their mean is 94.0368\%. On the 150 in-scope CLINC150 classes, probability fusion of one Laya specialist and three DeBERTa-v3-large classifiers achieves 98.0222\% (4,411/4,500). We describe the two configurations, reproduce the reference comparisons displayed in the accompanying evaluation explorer, and separate headline scores from evidence about improvement. Under a uniform BANKING77 evaluation path, the SAM mean exceeds the historical checkpoint by 0.0433 percentage points, with a shared-row 95\% interval of [-0.2273, +0.3247]. Thus, the continuation does not establish a reliable accuracy gain. The report documents strong observed task-specific performance, selection effects, and unresolved historical execution-path differences, rather than a new backbone or a unified algorithmic advantage.

\end{abstract}
\section{Introduction}
Intent classification requires mapping an utterance to a predefined outcome rather than generating an open-ended response. This distinction is particularly relevant to fine-grained benchmarks such as BANKING77 and CLINC150, which contain 77 banking intents and 150 in-scope intents, respectively \cite{ref2,ref3}. When classification is implemented through autoregressive text generation, the generated answer must be mapped back to an admissible label. A direct decision model instead produces a distribution over the permitted classes without decoding a textual answer. This motivates a prediction interface aligned with the task's finite output space, although the formulation alone does not guarantee better accuracy or lower computational cost than alternative implementations.

Laya provides a non-generative decision-model foundation \cite{ref1}, but adapting its flexible candidate interface to fixed-label benchmarks introduces an input-representation challenge. The original interface serializes questions, candidate descriptions, and utterance text into a shared encoder sequence, allowing the candidate inventory to compete with the utterance for the available token budget. Our archived CLINC150 audit identified incomplete candidate coverage under the evaluated input configuration. To address this issue, DeepOpen retains the Laya encoder while replacing the flexible candidate-scoring interface with task-specific classification heads. The encoder processes the utterance without repeatedly serializing the entire label inventory, and the output head represents every class explicitly. This adaptation trades arbitrary candidate flexibility for a fixed-label decision interface; it does not introduce a new backbone or preserve all capabilities of the original typed-decision model.

Building on this adaptation, we investigate two complementary routes to task-specific performance. For BANKING77, we apply sharpness-aware minimization (SAM) to continue training a single classifier, testing whether perturbation-aware optimization improves the existing decision boundary without increasing inference-time parameter count \cite{ref5}. For CLINC150, we combine an R-Drop-trained Laya specialist with three DeBERTa-v3-large classifiers through probability fusion, testing whether their predictions provide complementary information \cite{ref4,ref6}. These interventions address different questions: SAM modifies optimization, whereas fusion combines multiple predictors at inference. We therefore evaluate them as separate system configurations and distinguish their effects from the fixed-label adaptation that resolves candidate-coverage constraints.

The resulting systems achieve observed test accuracies of 94.1234\% on BANKING77 and 98.0222\% on in-scope CLINC150 \cite{ref7}. The BANKING77 result is the best observed among three SAM seeds and uses approximately 396M parameters, compared with the nominal 7B Llama 2 reference; the CLINC150 result uses a four-encoder ensemble totaling approximately 1.697B parameters \cite{ref11,ref14}. Our contributions are the documented adaptation of a flexible decision-model interface to fixed-label intent classification, the evaluation of task-specific optimization and fusion strategies, and an analysis connecting accuracy with parameter footprint and reproducibility. The experiments distinguish headline performance from evidence of improvement: the BANKING77 mean gain remains statistically inconclusive, while the CLINC150 fusion results must be interpreted alongside their additional inference resources.

\section{Methodology}
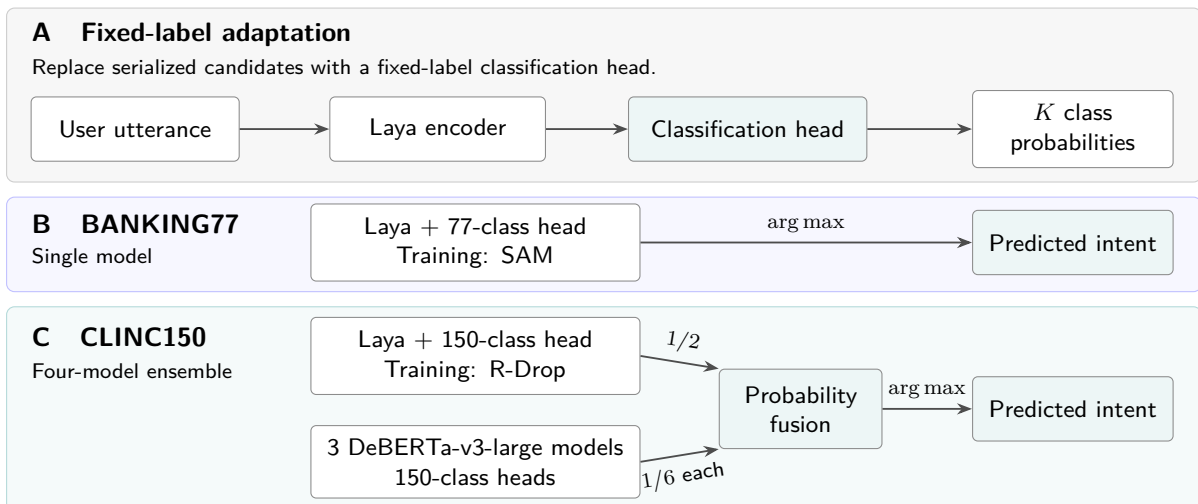
\begin{figure}[htbp]
\centering
\begin{tikzpicture}[
  x=1cm,y=1cm,
  font=\sffamily\fontsize{9}{11}\selectfont,
  box/.style={draw=black!45,rounded corners=2pt,fill=white,
    align=center,minimum height=0.85cm,inner sep=5pt},
  flow/.style={-{Stealth[length=2mm]},line width=0.7pt,draw=black!70},
  heading/.style={anchor=west,font=\sffamily\bfseries\fontsize{10}{12}\selectfont},
  note/.style={font=\sffamily\fontsize{8}{10}\selectfont,align=center}
]
\filldraw[fill=black!3,draw=black!20,rounded corners=3pt]
  (-0.2,0.35) rectangle (15.6,-1.95);
\filldraw[fill=blue!3,draw=blue!25,rounded corners=3pt]
  (-0.2,-2.15) rectangle (15.6,-3.4);
\filldraw[fill=teal!4,draw=teal!30,rounded corners=3pt]
  (-0.2,-3.6) rectangle (15.6,-6.3);
\node[heading] at (0,0) {A\quad Fixed-label adaptation};
\node[note,anchor=west] at (0,-0.45)
  {Replace serialized candidates with a fixed-label classification head.};
\node[box,text width=2.4cm] (input) at (1.5,-1.25) {User utterance};
\node[box,text width=2.5cm] (encoder) at (5.5,-1.25) {Laya encoder};
\node[box,text width=2.8cm,fill=teal!7] (head) at (9.6,-1.25) {Classification head};
\node[box,text width=2.3cm] (output) at (13.9,-1.25) {$K$ class probabilities};
\draw[flow] (input) -- (encoder);
\draw[flow] (encoder) -- (head);
\draw[flow] (head) -- (output);

\node[heading] at (0,-2.5) {B\quad BANKING77};
\node[note,anchor=west] at (0,-2.95) {Single model};
\node[box,text width=4.0cm] (bank) at (6.0,-2.75)
  {Laya + 77-class head\\{\footnotesize Training: SAM}};
\node[box,text width=2.3cm,fill=teal!7] (bankout) at (13.9,-2.75)
  {Predicted intent};
\draw[flow] (bank) -- node[above,note] {$\arg\max$} (bankout);

\node[heading] at (0,-4.0) {C\quad CLINC150};
\node[note,anchor=west] at (0,-4.45) {Four-model ensemble};
\node[box,text width=4.0cm] (laya) at (6.0,-4.25)
  {Laya + 150-class head\\{\footnotesize Training: R-Drop}};
\node[box,text width=4.0cm] (deberta) at (6.0,-5.65)
  {3 DeBERTa-v3-large models\\{\footnotesize 150-class heads}};
\node[box,text width=1.8cm,fill=teal!7,minimum height=1.05cm] (fusion) at (10.3,-4.95)
  {Probability\\fusion};
\node[box,text width=2.3cm,fill=teal!7] (clincout) at (13.9,-4.95)
  {Predicted intent};
\draw[flow] (laya.east) -- node[above,sloped,note] {$1/2$} (fusion.north west);
\draw[flow] (deberta.east) -- node[below,sloped,note] {$1/6$ each} (fusion.south west);
\draw[flow] (fusion) -- node[above,note] {$\arg\max$} (clincout);
\end{tikzpicture}
\caption{Overview of deepopen. (A) Fixed-label adaptation removes candidate descriptions from the input. The two tasks use separately trained models: (B) a single SAM-trained model for BANKING77 and (C) weighted probability fusion for CLINC150. SAM and R-Drop are used only during training.}
\label{fig:deepopen_overview}
\end{figure}

Figure~\ref{fig:deepopen_overview} presents the overall design
of DeepOpen. Panel~A shows the adaptation from Laya's flexible
candidate interface to fixed-label intent classification.
Panels~B and~C instantiate this design with separately trained
systems: a single classifier continued with sharpness-aware
minimization (SAM) for BANKING77, and probability fusion of
four classifiers for CLINC150. SAM and R-Drop affect training,
whereas the arrows in Panels~B and~C describe inference.

\subsection{SAM-Based Optimization for BANKING77}
The BANKING77 system uses a Laya-based English encoder and a 77-class classification head. Its starting checkpoint was trained on all 10,003 official training examples, including a subset previously used for internal development. That subset therefore cannot serve as independent validation for this checkpoint.

All continuation candidates start from the same historical full-training weights. Three fixed seeds (13, 42, 87) are run for the control and SAM candidates. The recorded protocol uses two epochs, 626 updates, 10,003 training rows, encoder learning rate 2e-6, head learning rate 2e-5, and BF16 training. SAM uses rho = 0.05 \cite{ref5}. Fixed final checkpoints are evaluated without validation-based checkpoint selection. Selecting seed 13 as the headline is a separate, post-test display choice.

Evaluation uses eager FP32, scaled dot-product attention (SDPA), batch size 32, and a maximum length of 128 tokens for every continuation checkpoint and the historical reference. The headline model performs inference with one encoder; no seed ensemble is used.

\subsection{Probability Fusion for CLINC150}
The CLINC150 system combines a Laya-based English specialist with three DeBERTa-v3-large classifiers \cite{ref6}. The specialist uses masked mean pooling and a 150-class linear head; the historical selected configuration uses R-Drop training \cite{ref4} and seed 42. The DeBERTa members use seeds 13, 42, and 87. Each member uses its own tokenization, and probabilities must share the same 150-class label order before fusion.

Let pL denote the specialist probabilities and p13, p42, and p87 the DeBERTa probabilities. The final distribution is:

\begin{equation}p_{\mathrm{ensemble}}=\tfrac12 p_L+\tfrac16(p_{13}+p_{42}+p_{87}).\end{equation}
The predicted intent is argmax(pEnsemble). All four encoders are required at inference. The archived hybrid protocol evaluated Laya weights in \{0, 0.25, 0.5, 0.75, 1\}, using validation accuracy and then NLL for selection. The final system uses weight 0.5 and equal weighting across the three DeBERTa members; selection was frozen before this evaluation. Earlier rounds had already exposed benchmark test summaries, so the overall development history is not a fresh blind evaluation \cite{ref14}.

\subsection{Relationship to Existing Methods}
Laya supplies the starting English model \cite{ref1}. R-Drop \cite{ref4}, SAM \cite{ref5}, and DeBERTa-v3 \cite{ref6} are established components, not methods introduced here. The paper reports their task-specific use and observed system behavior. It does not isolate the contribution of each component through a common cross-dataset ablation.

\section{Experiment}
\subsection{Dataset for Intent Classification}
\begingroup\small\renewcommand{\arraystretch}{1.15}
\begin{longtable}{>{\raggedright\arraybackslash}p{\dimexpr 0.17708\textwidth-2\tabcolsep\relax}>{\raggedright\arraybackslash}p{\dimexpr 0.39583\textwidth-2\tabcolsep\relax}>{\raggedright\arraybackslash}p{\dimexpr 0.12500\textwidth-2\tabcolsep\relax}>{\raggedright\arraybackslash}p{\dimexpr 0.30208\textwidth-2\tabcolsep\relax}}
\caption{Dataset scope and headline aggregation.}\\
\toprule
\textbf{Dataset} & \textbf{Training / evaluation scope} & \textbf{Test size} & \textbf{Headline aggregation}\\\midrule\endfirsthead
\textbf{Dataset} & \textbf{Training / evaluation scope} & \textbf{Test size} & \textbf{Headline aggregation}\\\midrule\endhead
\bottomrule\endfoot
BANKING77 & 10,003 full-training examples; 77 intents & 3,080 & SAM seed 13; best observed of three\\
CLINC150 & 150 in-scope intents; original train/dev/test split & 4,500 & Four-encoder probability fusion\\
\end{longtable}
\endgroup
Accuracy is correct predictions divided by test examples. BANKING77 continuation uncertainty uses 10,000 shared-row bootstrap resamples of fixed checkpoints. These intervals exclude model-selection effects and uncertainty over the broader training-seed population. They do not turn the observed tests into a new independent confirmation set. The public protocol records no subsequent tuning following these tests.

\subsection{Benchmark Results}
\begingroup\small\renewcommand{\arraystretch}{1.15}
\begin{longtable}{>{\raggedright\arraybackslash}p{\dimexpr 0.34375\textwidth-2\tabcolsep\relax}>{\raggedright\arraybackslash}p{\dimexpr 0.20833\textwidth-2\tabcolsep\relax}>{\raggedright\arraybackslash}p{\dimexpr 0.17708\textwidth-2\tabcolsep\relax}>{\raggedright\arraybackslash}p{\dimexpr 0.27083\textwidth-2\tabcolsep\relax}}
\caption{Headline results, aligned with the evaluation explorer retrieved on 21 October 2026 \cite{ref7}.}\\
\toprule
\textbf{System} & \textbf{Correct / test} & \textbf{Accuracy (\%)} & \textbf{Interpretation}\\\midrule\endfirsthead
\textbf{System} & \textbf{Correct / test} & \textbf{Accuracy (\%)} & \textbf{Interpretation}\\\midrule\endhead
\bottomrule\endfoot
BANKING77 SAM seed 13 & 2,899 / 3,080 & 94.1234 & Best observed SAM seed\\
CLINC150 four-encoder ensemble & 4,411 / 4,500 & 98.0222 & In-scope ensemble result\\
\end{longtable}
\endgroup
The BANKING77 headline is one correct prediction above the archived 94.0909\% result, but the execution settings differ; Section 5 gives the matched-path continuation analysis. The CLINC150 archive also includes specialist and DeBERTa-only ensemble comparisons, which are reported with their resource costs in Section 4.2.

\subsection{Separate Benchmark and Model Size Comparisons}
Accuracies below retain the selected evaluation-explorer values \cite{ref7}. Model sizes include embeddings and all ensemble members where verified. M denotes one million parameters. NV means the size of the exact displayed result configuration was not verified; it does not mean zero. These contextual comparisons are not reruns under one shared data, selection, or compute protocol.

\begin{table}[htbp]
\centering
\small
\renewcommand{\arraystretch}{1.2}
\caption{BANKING77 test accuracy comparison.}
\label{tab:banking77_comparison}

\begin{tabular*}{\textwidth}{
    @{\extracolsep{\fill}}lrl@{}
}
\toprule
\textbf{Method} &
\textbf{Accuracy (\%)} &
\textbf{Inference configuration} \\
\midrule
SPACE (SPACE-3) &
94.94 &
Single Transformer + classification head \\

SPACE 2.0 (SPACE-2) &
94.77 &
Single encoder + classification head \\

Llama 2 7B (FT) &
94.35 &
Single decoder-only model \\

ConvFiT (RoBERTa-base) &
94.16 &
RoBERTa-base + cosine 1-NN \\

\textbf{DeepOpen (SAM)} &
\textbf{94.1234} &
Single encoder + classification head \\

PPTOD-base &
93.86 &
T5-base; intent-label generation \\
\bottomrule
\end{tabular*}


\end{table}

The BANKING77 classifier uses approximately 17.7 times fewer parameters than the nominal 7B reference, with accuracy 0.2266 percentage points lower. This supports a parameter-footprint comparison for that reference pair, not an equal-compute or latency claim. 

\begin{table}[htbp]
\centering
\small
\renewcommand{\arraystretch}{1.2}

\caption{CLINC150 in-scope test accuracy comparison.}
\label{tab:clinc150_comparison}

\begin{tabular*}{\textwidth}{
    @{\extracolsep{\fill}}lrl@{}
}
\toprule
\textbf{Method} &
\textbf{Accuracy (\%)} &
\textbf{Inference configuration} \\
\midrule
TOATOD-small &
98.45 &
T5-small + task adapter \\

\textbf{DeepOpen (ensemble)} &
\textbf{98.0222} &
Laya + 3 DeBERTa-v3-large models \\

TOATOD-base &
98.01 &
T5-base + task adapter \\

SPACE (SPACE-3) &
97.89 &
Single Transformer + classification head \\

SPACE 2.0 (SPACE-2) &
97.80 &
Single encoder + classification head \\
\bottomrule
\end{tabular*}

\medskip
\begin{minipage}{\textwidth}
\footnotesize
\textit{Notes.}
Results cover the 150 in-scope classes only.
The DeepOpen ensemble combines four encoders through
probability fusion.
Boldface identifies our system, not the best result.
\end{minipage}

\end{table}

The archived member counts are 394,935,446 for the Laya specialist and 434,165,910 for each DeBERTa classifier, totaling 1,697,433,176. The DeBERTa-only comparison reuses the same checkpoints selected within the hybrid validation procedure; it is not a separately optimized DeBERTa-only baseline. Adding Laya increases accuracy by 0.5778 percentage points over that component ensemble, while the final system exceeds the specialist by 0.2667 points (twelve net correct predictions).

The hybrid report records conditional paired 95\% bootstrap intervals of [0.2444, 0.9111] points against the DeBERTa-only ensemble and [0.0667, 0.4667] against the Laya specialist. These intervals condition on the selected weights and do not cover repeated research selection or the distribution of possible training runs. This revision carries forward those archived analyses; it does not present them as new independent experiments.

The deployed four-encoder weights total 6,789,898,104 bytes in the archived package. On an NVIDIA L20, the archived batch-1, padding-64, BF16 measurement with 20 warmup passes and 100 measured passes reports p50 49.411 ms and p95 49.632 ms for resident forward computation and probability fusion. Loading, tokenization, and network time are excluded. The specialist has an historical p50 of 10.104 ms under a similar protocol, but this is not a same-process paired speed test.

\subsection{Parameter Accounting and Interpretation}
The BANKING77 total is derived, not recounted from a newly loaded checkpoint. The shared encoder count is 394,935,446 minus the CLINC150 linear head (1024 + 1) x 150, giving 394,781,696. The archived BANKING77 configuration and the Transformers implementation \cite{ref14} add a bias-free 1024 x 1024 dense layer, 1024 LayerNorm weights, and a (1024 + 1) x 77 output head: 395,910,221 parameters. SAM adds no inference parameters. The evidence and calculation are saved in the companion parameter-accounting file.

Parameter count is a model-footprint measure, not a substitute for latency, peak memory, FLOPs, or energy. Sequence length, precision, batching, software kernels, and the number of model evaluations all affect deployment cost. The BANKING77 result supports a smaller parameter footprint than the 7B reference; the CLINC150 result illustrates an accuracy-resource trade-off from four-model fusion.

\section{BANKING77 Continuation and Execution Audit}
\begingroup\small\renewcommand{\arraystretch}{1.15}
\begin{longtable}{>{\raggedright\arraybackslash}p{\dimexpr 0.54167\textwidth-2\tabcolsep\relax}>{\raggedright\arraybackslash}p{\dimexpr 0.22917\textwidth-2\tabcolsep\relax}>{\raggedright\arraybackslash}p{\dimexpr 0.22917\textwidth-2\tabcolsep\relax}}
\caption{All reported continuation candidates under the uniform eager FP32 path.}\\
\toprule
\textbf{Checkpoint} & \textbf{Correct / 3,080} & \textbf{Accuracy (\%)}\\\midrule\endfirsthead
\textbf{Checkpoint} & \textbf{Correct / 3,080} & \textbf{Accuracy (\%)}\\\midrule\endhead
\bottomrule\endfoot
Historical checkpoint, uniform path & 2895 & 93.9935\\
Control seed 13 & 2895 & 93.9935\\
Control seed 42 & 2894 & 93.9610\\
Control seed 87 & 2892 & 93.8961\\
SAM seed 13 (headline) & 2899 & 94.1234\\
SAM seed 42 & 2896 & 94.0260\\
SAM seed 87 & 2894 & 93.9610\\
Control three-seed mean & -- & 93.9502\\
SAM three-seed mean & -- & 94.0368\\
\end{longtable}
\endgroup
\begingroup\small\renewcommand{\arraystretch}{1.15}
\begin{longtable}{>{\raggedright\arraybackslash}p{\dimexpr 0.47917\textwidth-2\tabcolsep\relax}>{\raggedright\arraybackslash}p{\dimexpr 0.16667\textwidth-2\tabcolsep\relax}>{\raggedright\arraybackslash}p{\dimexpr 0.35417\textwidth-2\tabcolsep\relax}}
\caption{Fixed-checkpoint retrospective differences. All reported intervals include zero.}\\
\toprule
\textbf{Comparison} & \textbf{Delta (pp)} & \textbf{Shared-row 95\% interval (pp)}\\\midrule\endfirsthead
\textbf{Comparison} & \textbf{Delta (pp)} & \textbf{Shared-row 95\% interval (pp)}\\\midrule\endhead
\bottomrule\endfoot
SAM seed 13 vs historical & +0.1299 & [-0.1623, +0.4545]\\
SAM mean vs historical & +0.0433 & [-0.2273, +0.3247]\\
SAM mean vs control mean & +0.0866 & [-0.1299, +0.3139]\\
\end{longtable}
\endgroup
The SAM mean is 94.0368\%, compared with 93.9935\% for the historical checkpoint under the same evaluation path. Its +0.0433-point mean difference has an interval spanning zero. The SAM-control mean difference is +0.0866 points, also with an interval spanning zero. Neither comparison establishes a reliable gain. The 94.1234\% headline additionally carries post-test best-seed selection effects that these intervals do not model.

\subsection{Historical Prediction Drift}
\begingroup\small\renewcommand{\arraystretch}{1.15}
\begin{longtable}{>{\raggedright\arraybackslash}p{\dimexpr 0.54167\textwidth-2\tabcolsep\relax}>{\raggedright\arraybackslash}p{\dimexpr 0.22917\textwidth-2\tabcolsep\relax}>{\raggedright\arraybackslash}p{\dimexpr 0.22917\textwidth-2\tabcolsep\relax}}
\caption{Distinct evaluation records for the same historical BANKING77 checkpoint.}\\
\toprule
\textbf{Execution record} & \textbf{Correct / 3,080} & \textbf{Accuracy (\%)}\\\midrule\endfirsthead
\textbf{Execution record} & \textbf{Correct / 3,080} & \textbf{Accuracy (\%)}\\\midrule\endhead
\bottomrule\endfoot
Archived historical predictions & 2,898 & 94.0909\\
Current replay of original BF16 script & 2,897 & 94.0584\\
Uniform eager FP32 evaluation & 2,895 & 93.9935\\
\end{longtable}
\endgroup
The audit reports matching data and weight hashes but unresolved exact prediction reproduction. The current original-script replay differs on three predicted labels from the archive. Eight audited eager conditions yield 2,895 correct predictions. Execution-path sensitivity is demonstrated, but the responsible operator and remaining historical drift have not been isolated. These records must remain separate; selecting a favorable execution path would not establish a robust training improvement.

\section{Reproducibility and Limitations}
The evaluation explorer and its linked JSON records \cite{ref7} provide the result source for this revision. The continuation protocol includes hashes for the historical model and all six continuation checkpoints, training configuration, precision, and the bootstrap scope. The downloaded page and three JSON files are retained with SHA-256 hashes in the companion source manifest. No new training or model inference was performed for this editorial revision.

 The archived hybrid report records a full 4,500-example rerun of the exported CLINC150 system with zero prediction differences. Separate historical package checks on two synthetic examples establish loading behavior only. Member configurations, parameter counts, and fusion selection are available in the archive; Historical hardware provenance and a uniformly rerun cross-model benchmark remain incomplete.

Several limits affect interpretation. The two tasks use different architectures and objectives, so their scores do not establish a common causal improvement. The BANKING77 headline is selected after test evaluation, while its mean gain remains uncertain. Its bootstrap intervals exclude selection and seed-population uncertainty. CLINC150 component comparisons condition on hybrid-selected checkpoints and repeated benchmark exposure. Some external reference model sizes remain unverified, and parameter counts do not establish compute-matched efficiency. Finally, pretraining contamination and semantic overlap have not been ruled out; in-scope accuracy does not establish out-of-scope, robustness, or deployment performance.

This report is scoped to two English benchmarks and should not be interpreted as a complete inventory of all project experiments or evidence of multilingual generalization. The public page is a result display, not itself proof of accepted leaderboard submission or verified delivery of every model package.

\section{Conclusion}
DeepOpen achieves an observed BANKING77 accuracy of 94.1234\% with a single SAM-continued classifier and CLINC150 in-scope accuracy of 98.0222\% with a four-encoder ensemble. These figures match the specified evaluation explorer. Reporting the complete BANKING77 continuation results and execution audit gives necessary context: the three-seed mean is 94.0368\%, and its gain over the uniform-path historical checkpoint remains statistically inconclusive. The systems provide documented task-specific results; stronger causal, efficiency, or generalization claims require additional controlled evidence.


\begin{thebibliography}{99}
\bibitem{ref1} B. Warner et al. 2025.
\href{https://aclanthology.org/2025.acl-long.127/}{
Smarter, Better, Faster, Longer: A Modern Bidirectional
Encoder for Fast, Memory Efficient, and Long Context
Finetuning and Inference}.
In \emph{Proceedings of the 63rd Annual Meeting of the
Association for Computational Linguistics}.

\bibitem{ref2} I. Casanueva, T. Temcinas, D. Gerz, M. Henderson, and I. Vulic. 2020.
\href{https://aclanthology.org/2020.nlp4convai-1.5/}{Efficient Intent Detection with Dual Sentence Encoders}.
In \emph{Proceedings of NLP4ConvAI}.

\bibitem{ref3} S. Larson et al. 2019.
\href{https://aclanthology.org/D19-1131/}{An Evaluation Dataset for Intent Classification and Out-of-Scope Prediction}.
In \emph{Proceedings of EMNLP-IJCNLP}.

\bibitem{ref4} X. Liang et al. 2021.
\href{https://arxiv.org/abs/2106.14448}{R-Drop: Regularized Dropout for Neural Networks}.
\emph{Advances in Neural Information Processing Systems}.

\bibitem{ref5} P. Foret, A. Kleiner, H. Mobahi, and B. Neyshabur. 2021.
\href{https://arxiv.org/abs/2010.01412}{Sharpness-Aware Minimization for Efficiently Improving Generalization}.
In \emph{International Conference on Learning Representations}.

\bibitem{ref6} P. He, J. Gao, and W. Chen. 2021.
\href{https://arxiv.org/abs/2111.09543}{DeBERTaV3: Improving DeBERTa using ELECTRA-Style Pre-Training with Gradient-Disentangled Embedding Sharing}.
\emph{arXiv preprint arXiv:2111.09543}.

\bibitem{ref7} DeepOpen authors. 2026.
\href{https://c41-laya-evaluation.static.hf.space/index.html?version=ensemble-2#comparison}{Evaluation Explorer: Benchmark Comparison}.
Online evaluation resource, version ensemble-2. 



\bibitem{ref10} N. Bang, J. Lee, and M.-W. Koo. 2023.
\href{https://aclanthology.org/2023.findings-acl.464/}{Task-Optimized Adapters for an End-to-End Task-Oriented Dialogue System}.
In \emph{Findings of the Association for Computational Linguistics: ACL 2023}.

\bibitem{ref11} Meta.
\href{https://huggingface.co/meta-llama/Llama-2-7b-hf}{Llama 2 7B Model Card}.
Hugging Face model documentation.

\bibitem{ref12} Microsoft.
\href{https://huggingface.co/microsoft/deberta-v3-large}{DeBERTa-v3-large Model Card}.
Hugging Face model documentation.


\bibitem{ref14} Hugging Face.
\href{https://github.com/huggingface/transformers/blob/v4.57.6/src/transformers/models/modernbert/modeling_modernbert.py}{Transformers: ModernBERT Implementation}.
Software source code, version 4.57.6.
\end{thebibliography}
\end{document}